\documentclass[sigconf,nonacm]{acmart}
\AtBeginDocument{%
  }

\renewcommand\footnotetextcopyrightpermission[1]{}
\usepackage{todonotes}
\usepackage{float}

\begin{document}

\title{STARIXNet: Multivariate and Multi-attribute Deep Learning Approach to Real-Time Resource Allocation in Cloud Platforms}

\author{Ahmed Abdulaal}
\authornote{Work done while at Walmart Global Tech; Now at Gap Inc.}
\email{ahmedr.abdulaal@gmail.com}
\orcid{0000-0002-1607-2514}
\affiliation{%
  \institution{Walmart Global Tech}
  \city{Sunnyvale}
  \state{California}
  \country{USA}
}

\author{Maruf Aytekin}
\email{ahmet.aytekin@walmart.com}
\affiliation{%
  \institution{Walmart Global Tech}
  \city{Sunnyvale}
  \state{California}
  \country{USA}
}

\author{Thilaga kumaran Srinivasan}
\authornote{Both authors contributed equally to this research.}
\email{thilagakumaran.srini@walmart.com}
\author{Tomer Lancewicki}
\authornotemark[2]
\email{tomer.lancewicki@walmart.com}
\affiliation{%
  \institution{Walmart Global Tech}
  \city{Sunnyvale}
  \state{California}
  \country{USA}
}

\renewcommand{\shortauthors}{Abdulaal, Ahmed, et al.}

\begin{abstract}
    Intelligent scaling of microservices in cloud platforms is crucial for mitigating escalating compute costs while avoiding service disruptions. Current solutions are limited to the univariate space, typically focusing on CPU usage alone to drive scaling decisions. Moreover, they address the problem as a purely forecasting task, focusing on prediction precision while neglecting the greater risks of underestimation and delays in system responsiveness. Alternative solutions are computationally complex, making them impractical for large-scale, real-time deployments. To address these challenges, we present STARIXNet, a lightweight neural network that guides resource allocation decisions in the multivariate space by capturing spatio-temporal relationships among 
    multiple system metrics.
    STARIXNet models multiple quasi-dependent attributes, in particular the (S)easonal, (T)emporal, (A)uto-(R)egressive (I)ntegrated, and e(X)ogenous patterns, then implements an aggregation policy to finalize scaling decisions, prioritizing service stability, followed by cost-efficiency, over raw forecast accuracy. We empirically demonstrate the performance of STARIXNet by benchmarking against existing solutions 
    in real-world settings. STARIXNet is deployed for critical production microservices at Walmart achieving tangible savings ranging from 10\% to 50\%, in addition to intangible benefits through improved service stability and customer experience.
\end{abstract}

\maketitle

\section{Introduction}

Cloud platforms have long relied on intelligent autoscaling to dynamically adjust resources in response to workload fluctuations. This elasticity is essential for controlling costs (by releasing idle capacity) while upholding service reliability under peak loads \cite{Islam2012Elasticity}. Traditionally, autoscaling strategies are categorized as reactive – adding or removing resources based on threshold rules – or proactive – forecasting future demand and scaling in advance \cite{lorido2014review}. Reactive mechanisms, such as scaling out when CPU usage exceeds a fixed threshold, are simple but often lag behind sudden load surges, leading to performance degradation or Service Level Agreement (SLA) violations \cite{Chen2018Autonomous}. Conversely, proactive approaches attempt to predict workload trends to preemptively provision capacity \cite{Islam2012Elasticity}. In theory, proactive scaling can improve responsiveness, but in practice, current implementations suffer notable limitations.

Most proactive autoscaling systems focus on univariate forecasts, typically tracking a single metric, such as CPU usage or request rate, as the sole indicator of workload \cite{lorido2014review}. This narrow view misses the complex multi-resource dynamics of modern cloud applications. For example, the performance of an application can be jointly influenced by CPU, memory, network, and I/O; scaling decisions based solely on CPU can become suboptimal if another resource becomes the bottleneck \cite{ahuja2025effective}. Relying on one metric can also make autoscaling brittle. For example, CPU-only policies may lead to over-provisioning in response to CPU spikes, even when memory or network resources remain underutilized, resulting in inefficient resource use \cite{lorido2014review}. Additionally, many forecast-based autoscaling solutions update on coarse intervals, such as 15-minute and even as high as hourly cadence \cite{luo2024integrating, hua2023kae}, in order to reduce computational costs. However, this significantly impedes responsiveness to rapid workload changes. However, by reducing prediction intervals or using high-frequency metrics to improve agility introduces substantial overhead on the monitoring and scaling infrastructure \cite{Islam2012Elasticity}. Thus, operators face a trade-off between responsiveness and system overhead or stability \cite{Islam2012Elasticity}. Indeed, naive designs reacting to every short-term fluctuation often lead to oscillations (thrashing), where resources are repeatedly added and removed, compromising stability and increasing operational costs. In summary, existing univariate forecasting approaches often struggle with slow responsiveness to change, limited view of system state, and costly volatility in scaling behavior.

This paper introduces STARIXNet, a light-weight spatio-temporal deep learning solution for real-time decision-making, in order to address the aforementioned shortcomings. Instead of monitoring a single signal, STARIXNet learns a joint model of multiple metrics, including CPU usage, memory, network throughput, and more. It learns their hidden correlations over time, enabling a holistic view of application load patterns. Furthermore, STARIXNet provides several forecast options, based on different time series signal attributes and solution objectives. By employing a lightweight Deep Neural Network (DNN) architecture that captures both long-term trends and short-term spikes, STARIXNet achieves reliable predictions without the heavy computational footprint of complex generative, recurrent, and/or attention-based model architectures. Moreover, STARIXNet moves beyond pure prediction accuracy by tightly integrating the forecasting module with a decision engine that implements an aggregation policy prioritizing stability over point precision. In principle, the system favors consistent, gradual adjustments informed by spatio-temporal context, rather than reactive oscillation to every minor forecast deviation. This policy explicitly balances performance and cost: It reacts promptly to maintain Service Level Agreements (SLAs) and meet Service Level Objectives (SLOs), while dampening unnecessary resource transitions, a trade-off often neglected by prior approaches.

The key contributions of this work are summarized as follows:

\textbf{Multivariate Lightweight Architecture:} We innovatively design a deep learning model that fuses multiple resource metrics, external features,  learns quasi-dependent attributes, and captures spatio-temporal patterns for multivariate workload forecasting. In contrast to heavy attention-based models that introduce significant complexity \cite{Han2024SOFTS}, our architecture is optimized for real-time operation with linear scalability in input dimensions. It learns cross-metric interactions to anticipate resource needs more accurately than single-metric predictors, while remaining efficient and deployable. This design uniquely supports updating decisions at high frequencies and decentralized deployments, potentially as part of microservice sidecar pattern or agentic AI solution in future work.

\textbf{Customizable Stability-First Scaling Policy:} Our solution employs client-customizable aggregations and decision-making policies, with a default policy emphasizing system stability and SLA compliance over naively chasing every predicted fluctuation. By smoothing and validating forecasting outputs prior to executing scale actions, STARIXNet avoids the rapid oscillations observed in many aggressive autoscaling solutions. This novel policy addresses the known responsiveness-reliability trade-off \cite{Islam2012Elasticity}, through ensuring that scaling decisions are robust against transient noise, thus reducing thrashing and long-term costs.

\textbf{Real-World Deployment at Scale:} We report on the implementation of STARIXNet in a large-scale cloud platform, handling hundreds of microservices across geographically distributed data centers in real time. The framework was deployed with minimal friction alongside existing orchestration systems, such as kubernetes, demonstrating its practical integration capabilities. To our knowledge, this is one of the first real-time multivariate deep learning autoscaling approaches successfully validated in production at such scale.

\textbf{Measurable Performance Improvements:} Through extensive experiments, live benchmarking with real traffic, and live A/B testing, we show that STARIXNet achieves significant gains over state-of-practice alternatives. In our post-deployment analysis, we noted cloud resource cost reductions ranging from 10\% to 50\%, while simultaneously lowering SLA violation rates in comparison to all of default rule-based reactive, alternative deep learning, statistical, and univariate solutions. It outperformed advanced baselines in metrics such as average response time and scaling efficiency. These improvements underscore the value of coordinated multivariate learning and a stability-aware policy in real-time resource management.

The remainder of this paper is structured as follows: Section~\ref{sec:b} summarizes our findings from relevant literature. Section~\ref{sec:m} describes our solution in higher depth and mathematical notation. Section~\ref{sec:e} discusses our experimental setups covering live benchmarking, client simulations, and post-client onboarding evaluations. Section~\ref{sec:rnl} summarizes the observed results and impacts from both experiments and live, onboarded, critical microservices, as well as learned lessons and practical implications. Finally, we conclude this work in Section~\ref{sec:c}.

\section{Background}\label{sec:b}

Early cloud autoscaling mechanisms predominantly used simple threshold-based rules or queuing theory formulas to trigger scaling actions. Commercial and open-source cloud platforms, such as AWS Auto Scaling and Kubernetes Horizontal Pod Autoscaling (HPA) typically allow users to set static upper and lower bounds on a metric like average CPU usage or request queue length. Resources are added or removed when the metric crosses these thresholds \cite{lorido2014review, mao2010auto}. Such rule-based autoscaling methods are straightforward and react in real time, but require careful tuning and often perform sub-optimally in dynamic environments \cite{lorido2014review}, due to their rather reactive than proactive response, and negligence of pod startup delays. Choosing appropriate thresholds and cool-down durations require expert knowledge of each application's workload patterns \cite{lorido2014review}.

Recent approaches to dynamic resource allocation, such as in \cite{Chen2023EfficiencyAO, chen2023efficiency, mello2017architecture}, employ distributed orchestration and consensus algorithms to ensure rapid failure recovery by focusing on stateless application management, although full microservice decomposition remains constrained by system limitations in these designs. The authors in \cite{balla2020adaptive} proposed an alternative to HPA, which first vertically optimizes the resources for a given pod, then uses indicators from a virtual environment to a adapt resource definitions and adjust horizontal scaling accordingly, without requiring user-assigned parameters. 
However, these solutions remain reactive and limited to tracking a singular workload signal, the CPU usage.

Recent advances in machine learning and AI have led to a paradigm shift from traditional rule-based resource allocation to intelligent, predictive frameworks. Time series analysis frameworks are among the pioneering works in addressing the reactive lag of threshold methods by predicting future request rates and proactively scaling accordingly, thus improving SLA adherence in comparison to reactive rules \cite{calheiros2015workload, li2011improving}. However, these methods often assume stationarity and homoscedasticity, and most remain univariate despite the prevalence of multidimensional workloads in modern services \cite{lorido2014review}. To overcome these limitations, researchers have explored more flexible machine learning models, including transformer- and attention-based architectures \cite{mao2010auto, ahuja2025effective}. 
A notable example is Adaptive Horizontal Pod Autoscaling (AHPA), which enhances Kubernetes autoscaling under fluctuating business demand by combining decomposition-based time series forecasting with performance modeling in Alibaba Cloud Container Services \cite{zhou2023ahpa}. Similarly, \cite{rubak2023machine} propose a machine learning-driven solution to the challenges of over- and under-provisioning in Kubernetes. Their approach leverages classical models, including linear regression, support vector machines, and Multi-Layer Perceptron (MLP) neural networks to predict resource requirements based on anticipated user demand, enhancing both service quality and cost efficiency beyond what standard HPA can offer. However, while achieving robustness during seasonality fluctuations, these solutions remain questionable in addressing inconsistent or irregular workload patterns.

Recurrent Neural Networks (RNNs), particularly Long Short-Term Memory (LSTM) and Gated Recurrent Unit (GRU) architectures, have become widely adopted for modeling temporal patterns \cite{lai2018modeling}. The work in \cite{prachitmutita2018auto} utilized LSTM–MLP hybrid models to forecast web traffic, Bi-directional LSTMs were used in \cite{yan2021hansel}, while \cite{ouhame2021efficient} proposed hybrid Convolutional Neural Network (CNN) and LSTM (CNN–LSTM) architectures to capture spatial and temporal dependencies in cloud workloads. However, these solutions have widely addressed univariate workload signals and are characterized by high computational complexities, limiting wide-scale adoption in real-time settings.

The authors in \cite{toka2020adaptive} employed multiple machine learning models in a short-term evaluation loop, allowing them to compete under a "winner-takes-all" fusion strategy. Their multi-step-ahead predictive scaling engine, evaluated through both simulations and real-world web traces, showcased improved adaptability and efficiency. They also introduced a compact management parameter to help balance resource provisioning with adherence to SLA. 

More recently, Graph Neural Networks (GNNs) were utilized in \cite{luo2024integrating}, where the authors introduced a novel spatial-temporal graph neural network that models dynamic microservice interactions, workload periodicities, and system states, in order to enhance resource prediction and efficiency in microservice-based applications. While their method achieves notable gains in accuracy and real-world resource savings by comprehensively modeling spatio-temporal dependencies, it prioritizes prediction precision over computational speed, which may limit responsiveness in latency-critical scenarios. Furthermore, forecast precision metrics in isolation are misleading for the purpose of evaluating microservices autoscaling solutions, where the risk of workload underestimation is critically more costly than the risks of overestimation.

Another line of research explored reinforcement learning (RL) for autoscaling \cite{agarwal2024deep, tesauro2007online}. Researchers in \cite{xu2020firm}, applied deep RL to learn scaling policies optimizing SLA-cost trade-offs, and in \cite{xue2022morpheus}, the authors proposed meta model-based RL for better generalization to unseen workloads. However, RL faces challenges in sample efficiency, stability, and deployment complexity, especially in non-stationary cloud environments \cite{xue2022morpheus}. 

While the aforementioned solutions target predictive autoscaling, they often suffer from architectural complexity, platform specificity, and lack of large-scale real-world validation \cite{lorido2014review, calheiros2015workload, lai2018modeling, ouhame2021efficient, yan2021hansel}. Furthermore, key research gaps remain in multivariate modeling, stability-aware scaling policies, real-time and high-frequency revisions, and operational deployment at scale \cite{lorido2014review}. 

STARIXNet addresses these challenges by introducing a robust, lightweight, real-time, multivariate deep learning framework with a stability-first policy, validated through large-scale deployment at Walmart. This work bridges the gap between academic research and practical and scalable autoscaling solutions.

\section{Methodology}\label{sec:m}
In this section, we present STARIXNet in detail. STARIXNet is a DNN architecture combining four encoder-decoder modules, which share input features while utilizing different structures to model various attributes of the multivariate time series. The modeled time series targets describe workload in the multivariate space, including metrics critical to driving scaling decisions, such as CPU usage, memory, network throughput, request rate, or any other combination of user-specified metrics simultaneously. The attributes decoded and predicted by these modules are described as follows:
\begin{description}
\item[\texttt{Seasonal}:] Forecasts reflecting a Fourier series representation of the target variables.
\item[\texttt{Temporal}:] Forecasts reflecting acute and repetitive seasonal patterns, such as those due to planned events (e.g. maintenance or stress tests).
\item[\texttt{AutoRegressive-Integrated}:] Forecasts from the sequence-to-sequence (seq2seq) modeling of the multivariate targets, reflecting the short-term trends and dynamics.
\item[\texttt{eXogenous}:] Forecasts based on the spatio-temporal dependencies with external features, such as system or business features.
\end{description}

We denoted the name STARIXNet by combining the aforementioned attributes and the word "Network". Moreover, exogenous modules can be replicated for independent sets of external features, allowing the number of modules to be greater than four. For example, a microservice's system metrics, such as latency, utilization, errors, and network metrics, can represent one group of features requiring one exogenous module, while business metrics, such as Orders Per Minute (OPM) and user sign-in rate can represent another group, and upstream microservices' traffic can be a third group requiring a third exogenous module.

We summarize the nomenclature adopted in this section in Table~\ref{tab:nomenclature}. For simplicity, we use \(\Theta\) and \(\acute{\Theta}\) as generic symbols of any DNN function, \(\boldsymbol{\Theta}\) and \(\boldsymbol{\acute{\Theta}}\) as the sets of these DNN functions, while omitting typically redundant parameters (e.g. ReLU, linear layer weights, biases, depth, etc.), with the exception of parameters particular to our methods. Similarly, we use \(K\), \(L\), and \(Z\) in common notation form, however, the actual values of these parameters are different for each module or layer described in the following subsections. 
\begin{table}
    \caption{Nomenclature}
    \label{tab:nomenclature}
    \begin{tabular}{lp{5.5cm}}
        \toprule
        Symbol  &   Description \\
        \midrule
        \(N\)   &   Number of modeled resource load variables.  \\
        \(t\)   &   Index of time.  \\
        \(I\), \(J\)    &   Input and output sequence lengths.  \\
        \(\Theta\ \in \boldsymbol{\Theta}\), \(\acute{\Theta}\ \in \boldsymbol{\acute{\Theta}}\)    &   Encoder and decoder DNNs.   \\
        \(Y_{t-I:t-1} \in \mathbb{R}^{I \times N}\) &   Input tensor of target variables.   \\
        \(\acute{Y}_{t+1:t+J, \acute{\Theta}} \in \mathbb{R}^{J \times N}\) &   \(\acute{\Theta}\)'s output tensor of target variables. \\
        \(T_{t+1:t+J} \in \mathbb{Z}^{J}\)  &   1-dimensional vector of output timestamps.  \\
        \(s \in S\) &   index of seasonality encoding dimension.    \\
        \(\boldsymbol{\alpha}\), \(\boldsymbol{\omega}\), \(\boldsymbol{\beta} \in \mathbb{R}^{S \times N}\)    &   Learnable weights symbolizing amplitudes, angular frequencies, and phase shifts.    \\
        \(\boldsymbol{\gamma} \in \mathbb{R}^{N}\)  &   Vector of vertical displacements (i.e., biases). \\
        \(E \in \mathbb{R}^{|V| \times d}\) &   Embedding matrix.   \\
        \(K \in \mathbb{R}^{k_{T'}}\)   &   Convolutional kernels of size \(k_{T'}\).   \\
        \(l \in L\) &   Index of hidden layer for repeating layers. \\
        \(Z^{(l)}\) &   The \(l\)th layer learned representation.   \\
        \(M^{r}\)   &   Dimensionality of external features set \(r\).  \\
        \(X_{t-I:t-1}^{r} \in \mathbb{R}^{I \times M^{r}}\) &   Input tensor for a set \(r\) of external features.  \\
        \(\mathscr{w}_{\acute{\theta}}\), \(q_{\acute{\theta}}\)    &   Loss function's weight and quantile for \(\acute{\Theta}\). \\
        \(C \in \mathbb{Z}^{N}\)    &   Pod resource profile for the \(N\) resources.   \\
        \(U \in \mathbb{R}^{N}\)    &   Utilization targets for the \(N\) resources.    \\
        \bottomrule
    \end{tabular}
\end{table}

\begin{figure*}[t]
  \centering
  \includegraphics[width=\linewidth]{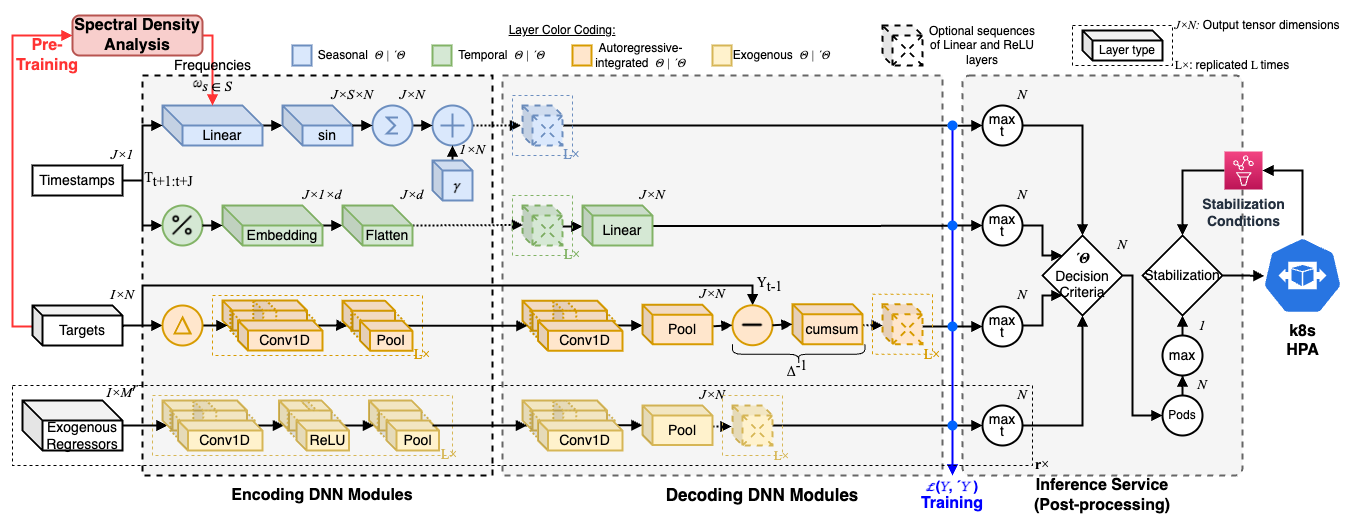}
  \caption{STARIXNet training and inference framework.}
  \Description{End-to-end architecture of the STARIXnet DNN with inference step.}
  \label{fig:e2e}
\end{figure*}

We illustrate the framework of STARIXNet schematically in Fig.~\ref{fig:e2e} and in the following subsections, we explain and present the mathematical formulation of the DNN's modules.

\subsection{Decoding the Seasonal Representation}\label{ssec:seasonal}
We designed a module in STARIXNet that learns the Fourier representation of the \(2\)-dimensional \(Y\), by encoding the 1-dimensional epoch timestamps, \(T_{t+1:t+J}\), using \(S\) linear layers with sinusoid activations, where optimized weights represent \(\boldsymbol{\alpha}\), \(\boldsymbol{\omega}\), and \(\boldsymbol{\beta}\), and reconstructs the future sequence, \(Y_{t+1:t+J}\), accordingly. We formulate this module as follows:
\begin{equation}\label{eq:seasonal1}
    \acute{Y}_{t+1:t+J, \acute{\Theta}} = \acute{\Theta} (\Theta (\boldsymbol{\alpha}, \boldsymbol{\omega}, \boldsymbol{\beta}, T) + \boldsymbol{\gamma})
\end{equation}
\begin{equation}\label{eq:seasonal2}
    \Theta  (\boldsymbol{\alpha}, \boldsymbol{\omega}, \boldsymbol{\beta}, T) = \sum_{s \in S} \boldsymbol{\alpha}_s \circ \sin(\boldsymbol{\omega}_s T_{t+1:t+J} + \boldsymbol{\beta}_s)
\end{equation}

Although the DNN, \(\Theta\), can independently learn appropriate weights for \(\boldsymbol{\omega}\) through backpropagation, we leverage prior knowledge of the modeled targets to initialize and freeze \(\boldsymbol{\omega}\), thus enhancing training efficiency. These priors include values representing daily and weekly frequencies (in radians), as well as values obtained from spectral analysis with Fast-Fourier Transform (FFT) \cite{welch2003use} on \(Y\) prior to training the DNN, similar to the approach proposed in \cite{abdulaal2021practical}.

The outputs of this module serve to represent the smoothed-periodic patterns while disregarding real-time deviations in other features. Consequently, it adds a layer of resilience to the inference service because it does not depend on the accuracy or consistency of any external metrics.

\subsection{Decoding the Temporal Representation}\label{ssec:temporal}
We designed another module that learns a more-discrete, periodic, representation of the \(2\)-dimensional \(Y\), by encoding the \(1\)-dimensional epoch timestamps, \(T_{t+1:t+J}\), using embedding learning with \(E\) weights matrix, such that the epoch timestamps are first transformed into a finite vocabulary set of size \(|V|\) using modulo operation, and mapped into an embeddings vector of  length \(d\), then reconstructs the future sequence, \(Y_{t+1:t+J}\), accordingly. We formulate this DNN module as follows:
\begin{equation}\label{eq:temporal1}
    \acute{Y}_{t+1:t+J, \acute{\Theta}} = \acute{\Theta} (\Theta (E, T))
\end{equation}
\begin{equation}\label{eq:temporal2}
    \Theta (E, T) = E[T_{t+1:t+J} \% |V|]
\end{equation}

The forecast outputs of this module serve to represent the discrete periodic pattern, while allowing for acute deviations, such as those due to predictable events (e.g., maintenance, stress tests, or promotional events). Similarly to the seasonal representation module, this module disregards real-time deviations in other features, thus adding an additional layer of resilience to the service.

\subsection{Autoregressive-Integrated Pattern Learning}\label{ssec:autoregressive}
Inspired by popular statistical state-space models such as ARIMA \cite{box2015time, calheiros2014workload}, we covered the autoregressive behavior in the DNN by encoding the multivariate input \(Y_{t-1:t-I, \acute{\Theta}}\) using matrix differencing and CNN operations, aiming to capture short-term spatio-temporal dependencies. The decoder layers then reconstruct the future sequence, \(Y_{t+1:t+J}\) from the feature maps. We formulate this DNN module as follows:
\begin{equation}\label{eq:autoregressive1}
    \acute{Y}_{t+1:t+J, \acute{\Theta}} = \Delta^{-1}\acute{\Theta} (\Theta (K, Y))
\end{equation}
\begin{equation}\label{eq:autoregressive2}
    \Theta (K, Y) = \{Pool(Conv1D(K^{(l)}, Z^{(l-1)}))\}    \qquad  \forall l \in L
\end{equation}
\begin{equation}\label{eq:autoregressive3}
    Z^{(l-1)} = \Delta Y_{t-I:t-1}    \qquad  \forall l=1
\end{equation}

This module serves to capture and forecast short-term, non-periodic, fluctuations, specifically with respect to steep upward or downward shifts in trend, such as those resulting from traffic failover across regions, DDoS attacks, social or promotional events, and many other factors. 

\subsection{Modeling Exogenous Features Influence}\label{ssec:exogenous}
Moreover, we include a group of DNN modules to model \(Y_{t+1:t+J}\) as functions of unique sets of exogenous regressors, \(X_{t-I:t-1}^{r}\). We encoded the past sequence \(X_{t-I:t-1}^{r}\) using CNN operations to capture spatio-temporal dependencies, then reconstructed the future sequence \(Y_{t+1:t+J}\) from the feature maps. We formulate the \(r\)th DNN module as follows:
\begin{equation}\label{eq:exogenous1}
    \acute{Y}_{t+1:t+J, \acute{\Theta}} = \acute{\Theta} (\Theta (K, X^{r}))
\end{equation}
\begin{equation}\label{eq:exogenous2}
    \Theta (K, X^{r}) = \{Pool(ReLU(Conv1D(K^{(l)}, Z^{(l-1)}))\}    \qquad  \forall l \in L
\end{equation}
\begin{equation}\label{eq:exogenous3}
    Z^{(l-1)} = X_{t-I:t-1}^{r}    \qquad  \forall l=1
\end{equation}

These modules serve to model and forecast potential fluctuations in the resource targets, in response to recent deviations in external groups of regressors, including system metrics, business metrics, or other interdependent microservices' traffic. For example, spikes or dips in top-of-the-funnel traffic features may indicate a potential change in the modeled microservice's load levels. In another example, an upward or downward trend in business metrics, such as the OPM or sign-in rate, may propagate to the modeled microservice's load levels. These modules aim to model such behaviors and more.

\subsection{Multi-Objective Loss Optimization}\label{ssec:loss}
One advantage of our DNN design, which outputs separate multivariate forecasts covering different attributes of the multivariate workload time series, is the enhanced user-flexibility of utilizing diverse loss functions or loss function parameters, tailored to the varying preferences of different microservices' owners. For example, a more risk-averse client might assign a higher penalty to underestimation than to overestimation when optimizing the loss functions for the seasonal and temporal forecasts. In contrast, a more risk-tolerant client may prefer more-balanced loss function parameters across all outputs. In another example, owners of microservices characterized by a high degree of noise, such as frequent momentary CPU spikes, may prefer reduced sensitivity in the autoregressive or some of the exogenous forecasts, in reaction to such noise. Additionally, users have the flexibility to adjust the weighting of the objectives, allowing for deviations from a strictly balanced multi-objective configuration. Accordingly, we provided, as a default configuration, a multi-objective quantile loss function for training the DNN, due to the flexibility of customizing quantile values for each \(\acute{\theta}\) output and user. We express this multi-objective loss as follows:
\begin{equation}\label{eq:loss}
    \begin{split}
        \mathscr{L}(Y,\acute{Y}) = \sum_{t} \sum_{\acute{\theta} \in \boldsymbol{\acute{\theta}}}  \mathscr{w}_{\acute{\theta}} \bigg( max \big[ q_{\acute{\theta}}(\acute{Y}_{t+1:t+J, \acute{\theta}} - Y_{t+1:t+J}),\\
        (q_{\acute{\theta}}-1)(\acute{Y}_{t+1:t+J, \acute{\theta}} - Y_{t+1:t+J})\big]
        \bigg)
    \end{split}
\end{equation}
where by default, we set \(\mathscr{w}_{\acute{\theta}}=1\) and \(q_{\acute{\theta}}=0.5\) for every \(\acute{\theta}\) unless the client requests a customization or depending on the unique characteristics of the onboarded microservice.

\subsection{Inference and Dynamic Scaling Policies}\label{ssec:inference}
In previous subsections, we discussed STARIXNet's architecture, parameters, and training scheme. In this subsection we briefly demonstrate basic inference and post-processing steps for translating STARIXNet's output into real-time scaling decisions. However, we must note that the exact implementation at Walmart involves many additional engineering and processing steps that are beyond the scope of this paper.

\subsubsection{Aggregation along the \(t\)-Dimension}\label{sssec:maxt}
\begin{figure}[h]
  \centering
  \includegraphics[width=\linewidth]{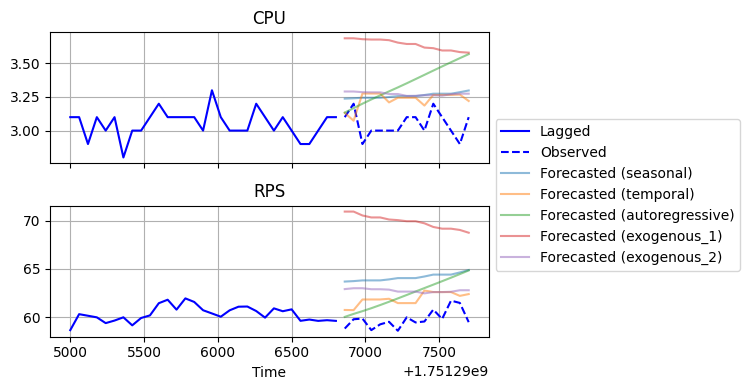}
  \caption{Input (lagged) and output (forecasted) targets.}
  \Description{Sample plotted input:output workload targets.}
  \label{fig:seq2seq}
\end{figure}

Another operational advantage of our approach, distinguishing it from prior reactive or point-based forecast solutions, is our utilization of a seq2seq framework, in order to achieve improved resilience and minimize the risk of delayed responsiveness. This is particularly critical in HPA problems, where pod start-up time and load-rebalancing delays, along with risks such as missing, delayed, or failed input signals, can significantly impact responsiveness or result in underestimation risks. Therefore, while inference executes at a near-real-time cadence, such as every 30 seconds, the forecasts obtained from each run instance span an extended horizon of length \(J\). For example, Fig.~\ref{fig:seq2seq} shows a single run instance's raw inputs and forecast outputs from a STARIXNet DNN with 5 modules and multivariate workload targets of size 2. The figure demonstrates that each update yields 15-steps ahead forecasts with different suggestions. Accordingly, we achieve risk minimization through maximum aggregation along the \(t\)-dimension:
\begin{equation}\label{eq:maxt}
    \acute{Y}_{t, \acute{\theta}} = \max_{t}(\acute{Y}_{t+1:t+J, \acute{\theta}})
\end{equation}

Furthermore, during both training and inference, we drop the signal values from the current runtime step, \(t=0\), as reflected in the aforementioned nomenclature. This design choice minimizes the risk of processing incomplete or partially available data. In other words, at time \(t=0\), STARIXNet takes inputs from the period \([t-I:t-1]\) and yields forecasts for the period \([t+1:t+J]\), deliberately ignoring potentially noisy readings obtained from \(t=0\).

\subsubsection{Aggregation along the \(\acute{\theta}\)-Dimension}\label{sssec:maxth}
The next step in post-processing involves combining the different types of forecasts. In this step, different users may elect to apply custom aggregation logic, as we had previously indicated. For example, a \(q\)-th quantile aggregation with a high \(q\) may suit risk-averse users. Alternatively, users may set custom rules or triggers, potentially guided by additional signals, to situationally switch between the different forecast types. However, the details of such strategies are beyond the scope of this paper. For simplicity, we present a \(q\)-th quantile aggregation method:
\begin{equation}\label{eq:quantile}
    \acute{Y}_t = \underset{\acute{\theta} \in \boldsymbol{\acute{\theta}}}{quantile}(q, \acute{Y}_{t+1:t+J, \acute{\theta}})
\end{equation}
\begin{figure}[h]
  \centering
  \includegraphics[width=\linewidth]{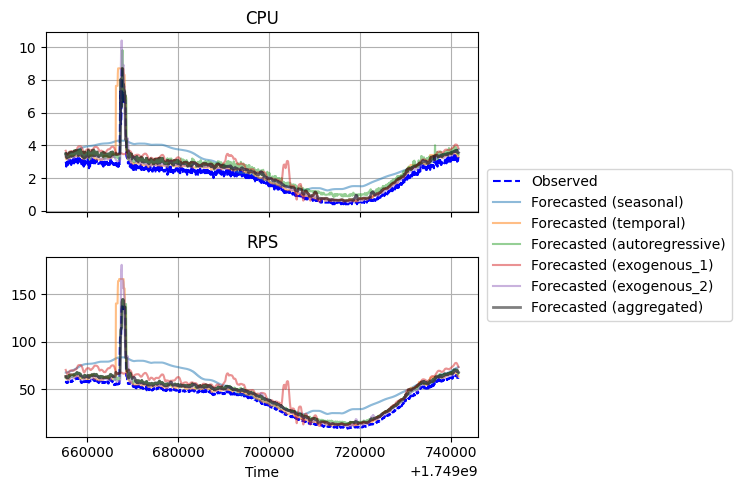}
  \caption{Median aggregation of the multivariate forecasts.}
  \Description{Sample plotted model output against the actual and the aggregated results.}
  \label{fig:median}
\end{figure}

In Fig.~\ref{fig:median} we show an example with \(q=0.5\) (i.e., median) across 5 forecast outputs for 2 workload metrics. The figure demonstrates the smoothing effect of this aggregation strategy, resulting in robustness to signal noise, without compromising responsiveness to meaningful fluctuations.

\subsubsection{Calculating the Number of Pods}\label{sssec:pods}
For computing the number of pods corresponding to each forecasted target value in \(Y_t\), we divide by the product of the per-pod resource capacity (i.e., the pod profile) and the corresponding utilization target. This results in \(N\) solutions for the pod count recommendations. We then take the maximum of the \(N\) recommendations, since a lower value would risk violating at least one of the utilization targets. We summarize this step mathematically, in vectorized form, as:
\begin{equation}\label{eq:pods}
    Pods_t = \max\big(\acute{Y}_t \oslash ({C \circ U})\big)
\end{equation}

\subsubsection{Applying Stabilization Measures and Driving HPA}\label{sssec:stabilization}
The final steps in post-processing include ensuring that fluctuations in \(Y_t\) over time would not risk system instability, then exposing the revised pod counts via Prometheus to drive HPA for the users' microservices using the {\verb|autoscaling/v2|} Kubernetes API and its custom metric scaling method. For the former step, we apply a short stabilization window to prevent subsequent downscaling events in a short period of time, as well as a separate, very short, stabilization window to avoid immediate upscaling events when a recent upscale has exceeded a pre-specified percentage threshold. The full implementation at Walmart involves many additional heuristics and engineering safeguards, which are beyond the scope of this paper.

\section{Experimental Setups}\label{sec:e}
In this section, we describe the setup of our benchmark study, conducted to evaluate the practical advantages and performance of STARIXNet against popular and State-Of-The-Art (SOTA) forecasting models. Moreover, we briefly describe the process of evaluating impacts and performance against the baseline, post-onboarding actual customer microservices.

\subsection{Live Benchmarking in Production Environment}\label{ssec:benchmark}

\subsubsection{System}\label{sssec:system}
To establish confidence, we avoided offline evaluations using artificial data or simulations. In lieu, we leveraged Walmart Cloud Native Platform (WCNP) to deploy identical clones of a Java-based microservice across multiple datacenter clusters and regions. We implemented a model-agnostic inference service and utilized a multi-instance deployment strategy to predict for each cloned microservices through a separate inference service instance, where each instance is configured to load a different forecasting model. This multi-instance deployment enabled us to collect and compare system performance metrics, such as CPU usage, memory usage, and inference latency (50th and 95th percentiles), corresponding to each loaded model type. For a visual illustration of the benchmark setup, readers can refer to Subsection~\ref{ssec:lbef} of the appendix. We note that this benchmark setup differs from our centralized inference service configuration used for actual customer microservices.

\subsubsection{Load}\label{sssec:load}
To generate the load, we have implemented a distributed traffic generator application that mirrors real traffic. Distributed instances of the traffic generator application generate identical requests across all cloned microservices for processing. This setup allows us to consistently simulate production-like load across services and easily test fail over and scaling scenarios. For fairness, all pods’ CPU, memory, and HPA bounds (i.e., minimum and maximum pod replicas) were set identically across all microservices and traffic generators. We also standardized the inference service architecture and CPU allocation, with exception of memory, since some benchmark models required significantly more RAM to load.

\subsubsection{Models}\label{sssec:models}
We benchmarked the live performance of STARIXNet against traditional univariate statistical models as well as multivariate deep learning and SOTA models. For statistical models, we selected \textbf{ARIMA} \cite{box2015time}, consistent with the experimentation baselines adopted in prior literature. Since ARIMA is a univariate model, we implemented a wrapper that trains separate ARIMA models for each load target and concatenates their outputs to match the multivariate format expected by our inference service. For multivariate DNN models, we implemented an \textbf{LSTM}-based RNN \cite{prachitmutita2018auto}, a probabilistic autoregressive RNN popularly known as \textbf{DeepAR} \cite{salinas2020deepar}, and the Temporal Fusion Transformer (\textbf{TFT}) \cite{lim2021temporal}, which incorporates SOTA transformers and attention mechanisms.

\subsubsection{Parameters and Configuration}\label{sssec:params}
One of the limitations to practically applying DNN solutions from literature is the costly and sensitive nature of hyperparameter tuning. This is especially challenging in decentralized settings, where thousands of microservices must each maintain models that are sensitive to frequent release cycles and evolving traffic patterns. Therefore, we sought a set of hyperparameter values that generalize well across most microservices, while maintaining reasonably light-weight models. Accordingly, we leveraged the built-in auto-tuning methods from the benchmark models' open-source Python packages, restricting the search to a modest range of values for DNN's depth, number of RNN layers, and dropout rates. It should be noted that for STARIXNet, we have not attempted any hyperparameter tuning during this benchmark study, and that it has fewer trainable parameters compared to the other DNN models. However, to ensure fairness, the following configuration values were held constant across all models, including STARIXNet: 30 (lagged steps) and 15 (forecast steps) for the encoder and decoder lengths, 256 for batch size, 100 maximum training epochs, and early termination patience of 10. Similarly, for the ARIMA models, we selected 30 for the autoregressive lag parameter. All models were trained on 1-minute interval time series of length 30,240. Furthermore, we included the exogenous features in all deep learning models where applicable, to ensure parity with STARIXNet.

\subsubsection{Forecast Targets and SLA}\label{sssec:sla}
In the multivariate context, we define three forecast targets as CPU usage, requests rate, and network throughput (i.e., network bytes received rate). The inference service uses the forecasts for these targets and computes the corresponding utilization metrics; CPU usage, Requests per pod, and network received rate per pod, then calculates the number of pods accordingly. The resulting values are aggregated and exposed as a custom scaling metric as explained in Subsection \ref{ssec:inference}. To maintain SLA, we set the utilization thresholds as follows: 25\% CPU usage, 10 requests per pod, and 25,000 network bytes received per pod. 

\subsection{Pre- and Post-Client Adoption Evaluations}\label{ssec:clienteval}
As with any product deployed at scale, we rely on a systematic evaluation and experimentation strategy to ensure reliability. Prior to onboarding a clients' microservice, we run simulations on historical data against the client's baseline, which is often a reactive CPU-based HPA or a pre-defined schedule-based scaling policy. Once the simulation results meet an agreed-upon acceptance criteria, we leverage the multiregion deployment of microservices to run quasi-A/B experimentation as part of a phased rollout plan. Our solution is initially adopted in one region and evaluated against the clients' baseline in the remaining regions for a few weeks. If the performance is satisfactory, we onboard a second region and repeat the process until the microservice is fully migrated to our solution.

\section{Results and Learnings}\label{sec:rnl}

\subsection{Live Benchmarking Results}\label{ssec:lbr}

\subsubsection{Inference System Performance}\label{sssec:isp}
The efficiency of real-time model inference at scale is crucial to maintaining timeliness, responsiveness, maintainability, and cost-effectiveness. This becomes particularly paramount when the inference component is integrated into the microservice deployment as a sidecar, alongside standard sidecar functionalities, such as load balancing, circuit breaking, dynamic configuration updates, and logging. While our currently adopted solution acts as an external microservice exposing the custom scaling metrics, our future vision is full decentralization, integrating the solution directly into the Kubernetes stack. Therefore, a practical solution must be lightweight in resources and latency. From the benchmark study, we conclude that our solution is superior in overall efficiency, as recorded in Table~\ref{tab:isp}, which reports average values of system performance metrics over a period of one day. We must note that we have not optimized STARIXNet or used quantization to enhance its performance. Surprisingly, ARIMA, despite being a traditional statistical solution, exhibited the highest resource consumption. We attribute this to the internal management required by the Kalman filter and the need to update state information before each forecast. While the recorded costs may appear trivial at benchmark study scale, the implications are significant at a large industrial scale and higher dimensional inputs. This aspect of inference performance benchmarking sets our work apart from prior literature, which often either overlooks inference system performance or focuses on non-real-time solutions.
\begin{table}
    \caption{Inference Service System Performance}
    \label{tab:isp}
    \begin{tabular}{lccc}
        \toprule
        Model       &   P50/95 Latency (ms)     &   Memory (MB) &   CPU (\%)        \\                           
        \midrule
        ARIMA       &   3.2/232                 &   17,406      &   3.1             \\
        DeepAR      &   3.2/4.8                 &   584         &   0.2             \\
        LSTM        &   \textbf{3}/\textbf{4.8} &   726         &   \textbf{0.1}    \\
        TFT         &   \textbf{3}/\textbf{4.8} &   779         &   0.2             \\
        STARIXNet   &   \textbf{3}/\textbf{4.8} &   \textbf{346}&   \textbf{0.1}    \\
        \bottomrule
    \end{tabular}
\end{table}

\subsubsection{Client-Side Impact}\label{sssec:csi}
As noted in early sections, we assert that the approaches used in some literature for evaluating models based on prediction accuracy metrics, such as mean squared errors, is potentially misleading in the context of autoscaling microservices. We know from experience that the adoption criteria for critical microservices are not driven by these metrics. Instead, clients prioritize operational performance indicators, including SLA compliance, system stability, alongside cost efficiency, particularly under irregular load patterns, peak periods, and during disaster recovery. In contrast, a high forecast accuracy does not guarantee business value. A single underestimation event can result in SLA violations with direct revenue loss and negative customer impact, far outweighing the impact of frequent overestimation. Therefore, we designed our solution to aggregate forecasts from both static and reactive model outputs, such as seasonality-based outputs and autoregressive outputs, thus preventing the end result from reaching dangerously low or unreasonably high predictions. Consequently, we evaluate performance based on the client's system health indicators, SLA/SLO-related metrics, and cost implications. We recorded these metrics from the benchmark experiment over a  period containing segments of irregular traffic patterns, and we summarize the findings in  Table~\ref{tab:cdsv} and Table~\ref{tab:cr1b}. Table~\ref{tab:cdsv} counts the instances in which traffic per replica, CPU usage, or latency thresholds were violated, while Table~\ref{tab:cr1b} compares compute cost and latency reductions relative to the model with the worst performance. The results demonstrate the effectiveness of our solution in balancing the competing goals of meeting SLAs and SLOs while reducing costs. For supplementary visualizations, readers may refer to Subsection~\ref{ssec:lbef} of the appendix.
\begin{table}
    \caption{Count of Daily SLA/SLO-related Metric Violations}
    \label{tab:cdsv}
    \begin{tabular}{lccccc}
        \toprule
        Metric  &   ARIMA   &   DeepAR  &   LSTM    &   TFT &   STARIXNet   \\  
        \midrule
        Load    &   1   &   2   &   3   &   \textbf{0}  &   \textbf{0}  \\
        CPU &   1   &   2   &   3   &   \textbf{0}  &   \textbf{0}  \\
        Latency &   \textbf{0}  &   \textbf{0}  &   1   &   \textbf{0}  &   \textbf{0}  \\
        \bottomrule
    \end{tabular}
\end{table}
\begin{table}
    \caption{Comparative Reductions with 100\% as Baseline}
    \label{tab:cr1b}
    \begin{tabular}{lccccc}
        \toprule
        Measure &   ARIMA   &   DeepAR  &   LSTM    &   TFT &   STARIXNet   \\
        \midrule
        Cost    &   42\%    &   41\%    &   100\%   &   46\%    &   \textbf{39\%}   \\
        Latency &   88\%    &   100\%   &   80\%    &   85\%    &   \textbf{64\%}   \\
        \bottomrule
    \end{tabular}
\end{table}

\subsection{Learnings from Onboarded Microservices}\label{ssec:fom}
The solution has been deployed at Walmart and is already managing autoscaling for many critical microservices, and continues to expand to additional services due to its reputable effectiveness. During phased rollouts, clients validated significant cost reductions, ranging from 10\% to 50\%, and in some instances noticeable improvements in system health metrics, based on cross-region comparisons. These findings were consistent with the pre-onboarding simulation results. It is important to note that the level of savings or improvements varies depending on each client's prior scaling strategy. We also note that the full end-to-end implementation involves various engineering aspects, guardrails, rules, and dependencies on other site reliability engineering tooling, which are beyond the scope of this paper. We share some anonymized dashboard screenshots from the monitoring of onboarded microservices in Section~\ref{sec:sf} of the appendix. One of the distinct advantages of our solution, acknowledged by clients, is its flexibility. It enables customizations and periodic modifications of inputs, thresholds, and aggregation strategies, to suit the various clients' needs, as well as to dynamically manage stability, risk, and cost trade-offs across different seasons and events. For supplementary visualizations, readers may refer to Subsection~\ref{ssec:ocmf} of the appendix.

\section{Conclusions}\label{sec:c}

Intelligent scaling of microservices in cloud platforms is central to managing compute costs and ensuring system stability. However, most existing solutions are limited in scope—treating the problem as a univariate forecasting task and optimizing for predictive accuracy at the expense of responsiveness, stability, and scalability. These shortcomings are exacerbated by the computational overhead of more advanced alternatives, which makes them unsuitable for large-scale, real-time environments.

To overcome these challenges, we introduced STARIXNet, a lightweight neural network designed for multivariate resource allocation decisions in dynamic cloud environments. Unlike conventional approaches, STARIXNet captures rich spatio-temporal dependencies across multiple quasi-dependent system metrics, including CPU usage, memory, request rate, and network throughput. By modeling Seasonal, Temporal, Auto-Regressive, Integrated, and eXogenous (STARIX) patterns, it formulates scaling decisions through a policy that explicitly prioritizes service stability, followed by cost-efficiency rather than raw forecast accuracy.

Empirical evaluations demonstrate the real-world efficacy of STARIXNet, deployed across critical production microservices at Walmart. The system achieved substantial infrastructure cost savings, ranging from 10\% to 50\%, while also improving service reliability and the end-user experience. These outcomes confirm the effectiveness of our approach and establish STARIXNet as a pragmatic, production-ready solution to intelligent autoscaling.

More broadly, our research contributes to the evolving landscape of machine learning for systems, building upon and extending recent developments in temporal modeling \cite{lim2021temporal}, state-aware workload prediction \cite{luo2024integrating}, and unified autoscaling frameworks \cite{zou2024optscaler, zhou2023ahpa}. The principles behind STARIXNet echo the shift towards combining deep learning, probabilistic reasoning \cite{salinas2020deepar}, and adaptive control policies for intelligent infrastructure management.

Additionally, our findings reinforce the need for scalable and interpretable models in production environments. While prior work has emphasized the importance of accuracy and complexity tradeoffs \cite{dixit2021machine}, STARIXNet demonstrates that it is possible to strike a balance—delivering performance gains while maintaining the simplicity and robustness necessary for real-time deployment.

In conclusion, STARIXNet represents a significant step forward in multivariate, policy-driven autoscaling, providing both a theoretical framework and a proven production-ready system that aligns with the operational goals of modern cloud infrastructure. We hope this work inspires further innovation at the intersection of AI, cloud systems, and operational excellence.

\bibliographystyle{unsrt}
\bibliography{arxiv_bibliography}

\appendix

\section{Reproducibility Information}\label{sec:ri}

\subsection{Code Access}

The initial release of the code, along with the licensing details, can be found at https://huggingface.co/aabdulaal/starixnet/tree/v1.0.0. To obtain access, please reach out to the corresponding author. The repository will be made publicly available once the current revision is finalized. Meanwhile, we include additional reproducibility information in the subsections below.

\subsection{Software Packages}

Independent of the client-facing production implementation, the live benchmark experiments deploy Python (slim-bullseye) runtime containers, installing the following dependencies:
\begin{itemize}
    \item Packages required by STARIXNet model:
    \begin{itemize}
        \item python 3.11.12
        \item joblib 7.8.0
        \item numpy 2.2.5
        \item pandas 2.2.3
        \item scipy 1.15.2
        \item torch 2.6.0 (CPU build)
        \item tqdm 4.67.1
    \end{itemize}
    \item Additional Packages required by benchmark models:
    \begin{itemize}
        \item lightning 2.5.1.post0
        \item pytorch-forecasting 1.3.0
        \item statsmodels 0.14.4
    \end{itemize}
    \item Additional packages installed for inference service:
    \begin{itemize}
        \item aiohttp 3.11.12
        \item aiosignal 1.3.2
        \item asyncio 3.4.3
        \item dataclasses-json 0.6.7
        \item dynaconf 3.2.11
        \item flask 3.1.0 (async)
        \item flask-caching 2.3.1
        \item prometheus-client 0.21.1
        \item tenacity 9.1.2
    \end{itemize}
\end{itemize}

We omit the listing of packages related to commercial storage solutions or internal tooling at Walmart to maintain compliance with policies.

\subsection{STARIXNet Parameter Configurations}
As mentioned in Subsection~\ref{ssec:benchmark}, we do not attempt separate hyperparameter tuning when training STARIXNet for new microservices or different sets of metrics. The per-module trainable parameters scale linearly in \(N\), while the exogenous modules scale linearly in \(NM^r\) for each \(r\). While there may be room to further improve precision through dedicated hyperparameter tuning, this would conflict with our goal of keeping the DNN as light-weight as possible. The general set of parameters we implement are as follows:
\begin{itemize}
    \item \(J = 15\)
    \item For seasonal module:
    \begin{itemize}
        \item \(S = 10\)
    \end{itemize}
    \item For temporal module:
    \begin{itemize}
        \item \(|V| = 10080\) and \(d = min(50, \frac{|V| + 1}{2}) = 50\)
        \item \(L = 1\)
    \end{itemize}
    \item For autoregressive-integrated module:
    \begin{itemize}
        \item \(k_{T'} = (N,  min(50, \frac{I + 1}{2}))\)
        \item \(L = 1\)
    \end{itemize}
    \item For any exogenous module:
    \begin{itemize}
        \item \(k_{T'} = (M^r,  min(50, \frac{I + 1}{2}))\)
        \item \(L = 2\)
    \end{itemize}
\end{itemize}

Therefore, the variables that may deviate by microservice are \(N\), \(M^r\), \(r\) and \(I\), where \(I=30\) was used for the benchmark study.


\section{Supplementary Figures}\label{sec:sf}

\subsection{Live Benchmark Experiment Figures}\label{ssec:lbef}

\begin{figure}[H]
  \centering
  \includegraphics[width=\linewidth]{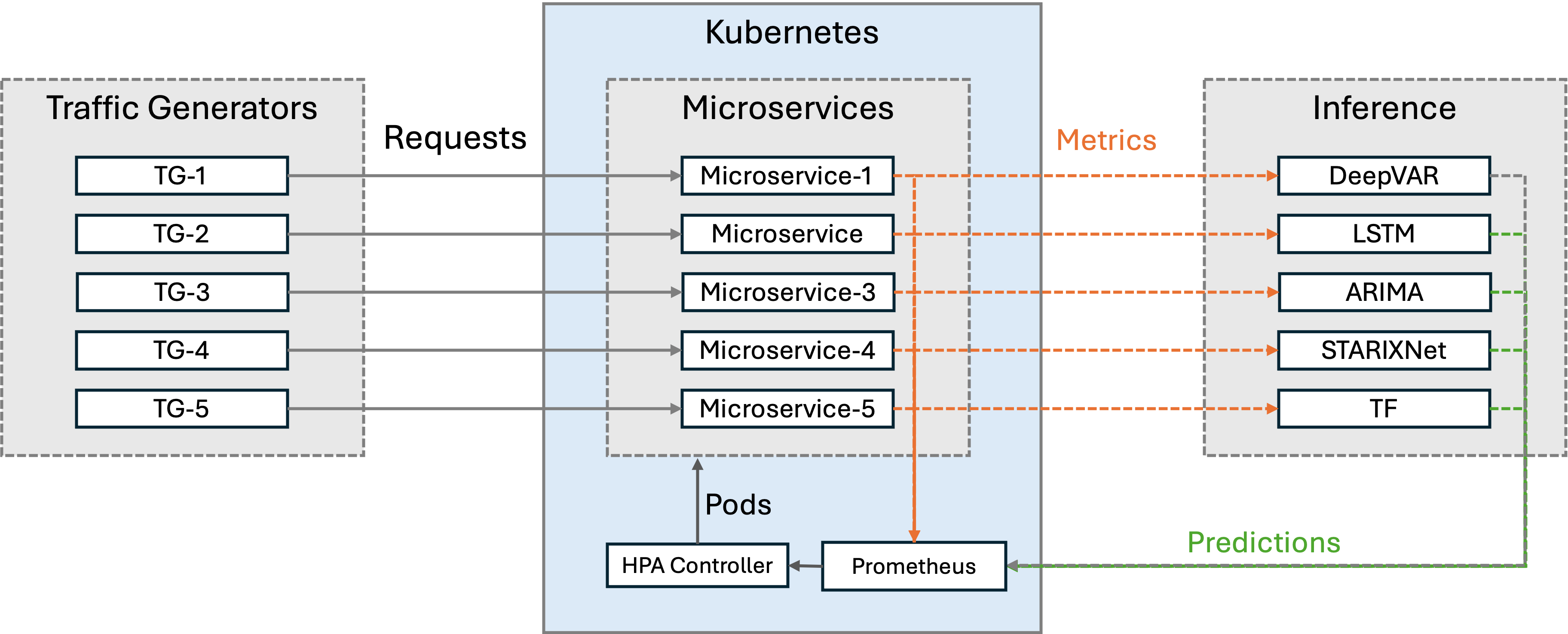}
  \caption{Experiment deployment architecture for STARIXNet load testing. Distributed instances of a traffic generator application mirror real traffic by sending identical requests to cloned microservice pods. All microservice pods share identical CPU and memory limits and HPA bounds (min/max replicas). Inference service pods use standardized CPU allocations, with memory adjusted per model requirements.}
  \Description{Block diagram showing distributed traffic generators sending load to cloned microservice pods instrumented with Prometheus sidecars; metrics flow into Prometheus, where the inference service reads them and feeds predictions back; the Kubernetes HPA controller then reads those predictions and adjusts microservice pod counts.}
  \label{fig:experiment_deployment_architecture}
\end{figure}

\begin{figure}[h!]
\centering
\includegraphics[width=\linewidth]{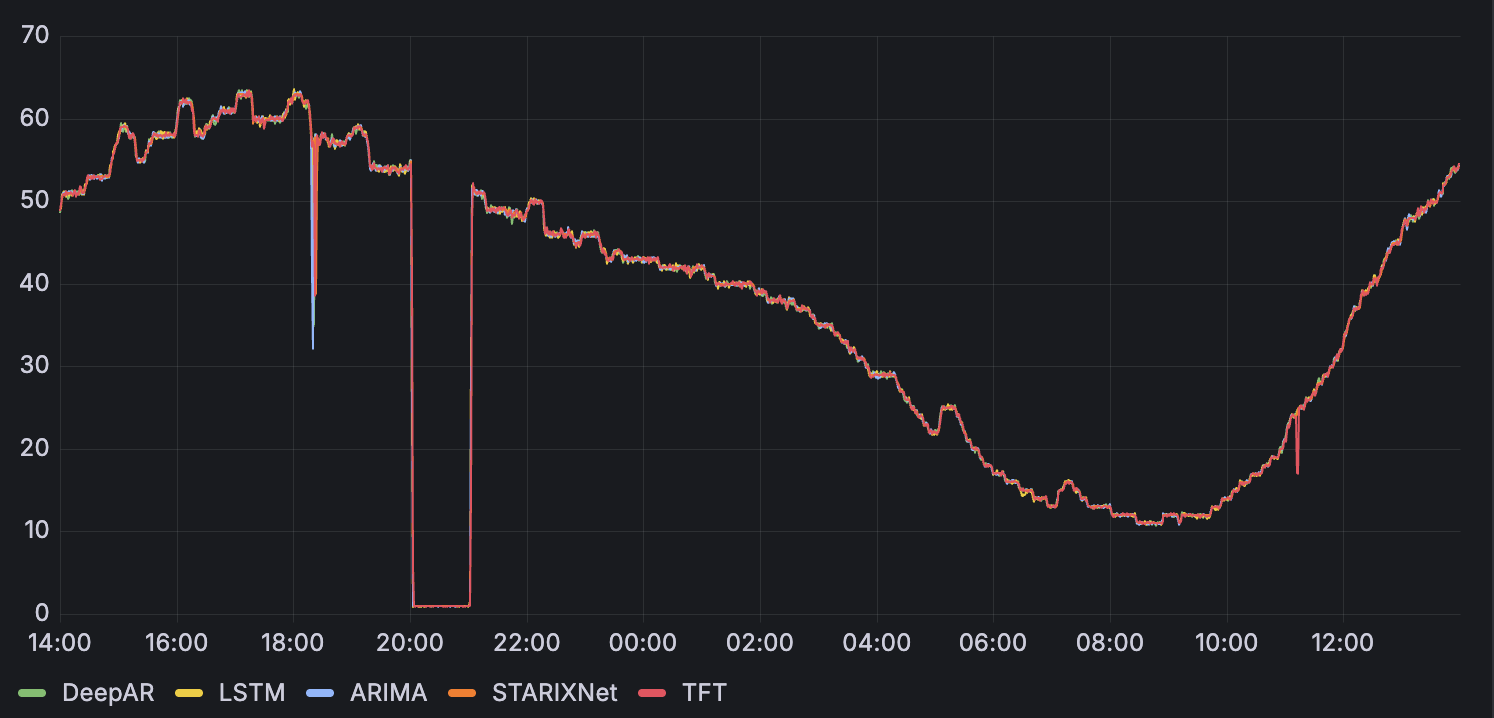}
\caption{Snapshot of the identical load pattern generated across all microservices. The distributed traffic generator sends synchronized requests to all cloned microservice pods, enabling consistent and fair comparison. The y-axis shows load in Requests Per Seconds (RPS); the x-axis shows the time. The traffic pattern includes steady phases, drops, and recoveries.}
\Description{Load generation accross all microservices.}
\label{fig:traffic_pattern}
\end{figure}

\begin{figure}[H]
\centering
\includegraphics[width=\linewidth]{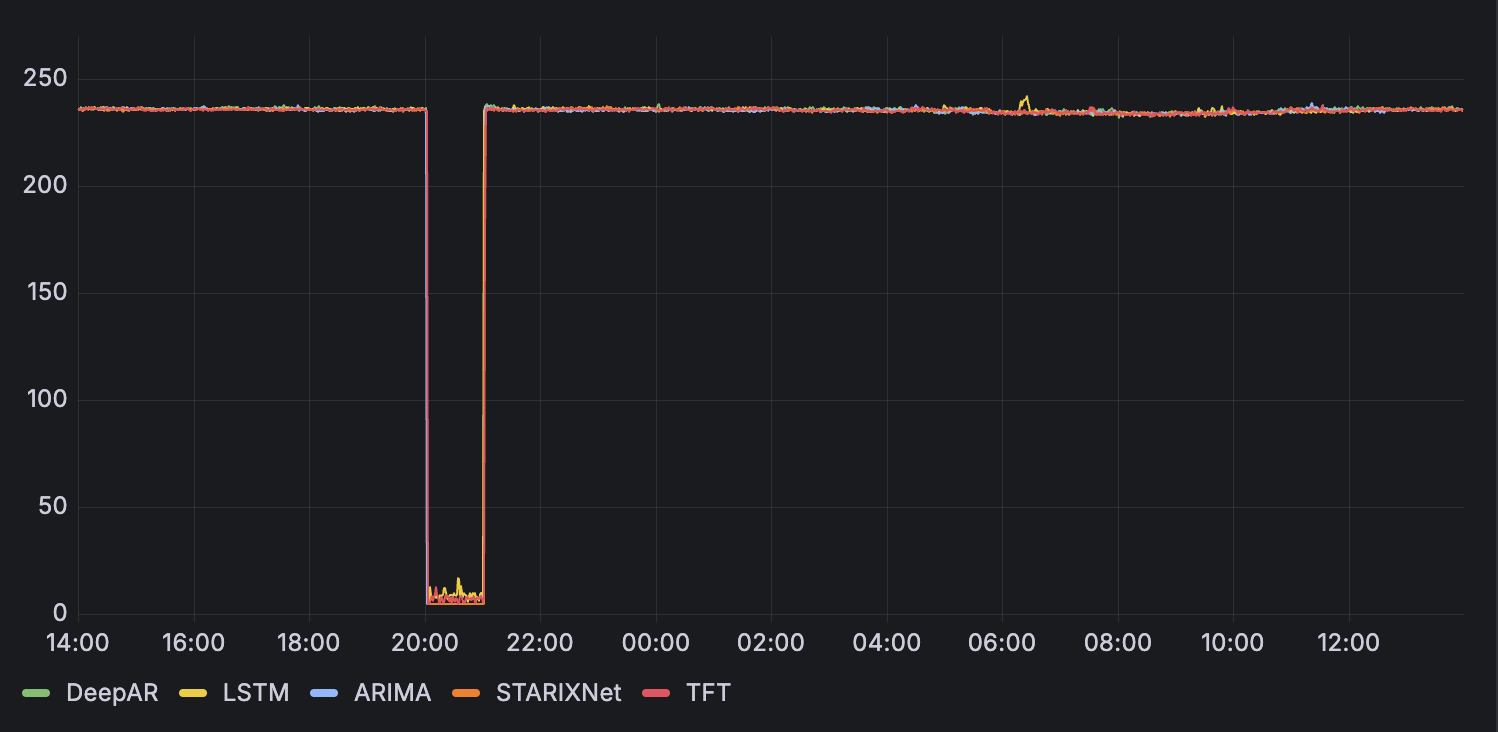}
\caption{Response latency (95th percentile) comparison across microservices during the evaluations. The y-axis shows latency in milliseconds; the x-axis shows the time. All models exhibit consistent latency patterns with notable drops during the periods of zero traffic load.}
\Description{Response latency (p95) for all microservices during the evaluations.}
\label{fig:latency_p95}
\end{figure}

\begin{figure}[H]
\centering
\includegraphics[width=\linewidth]{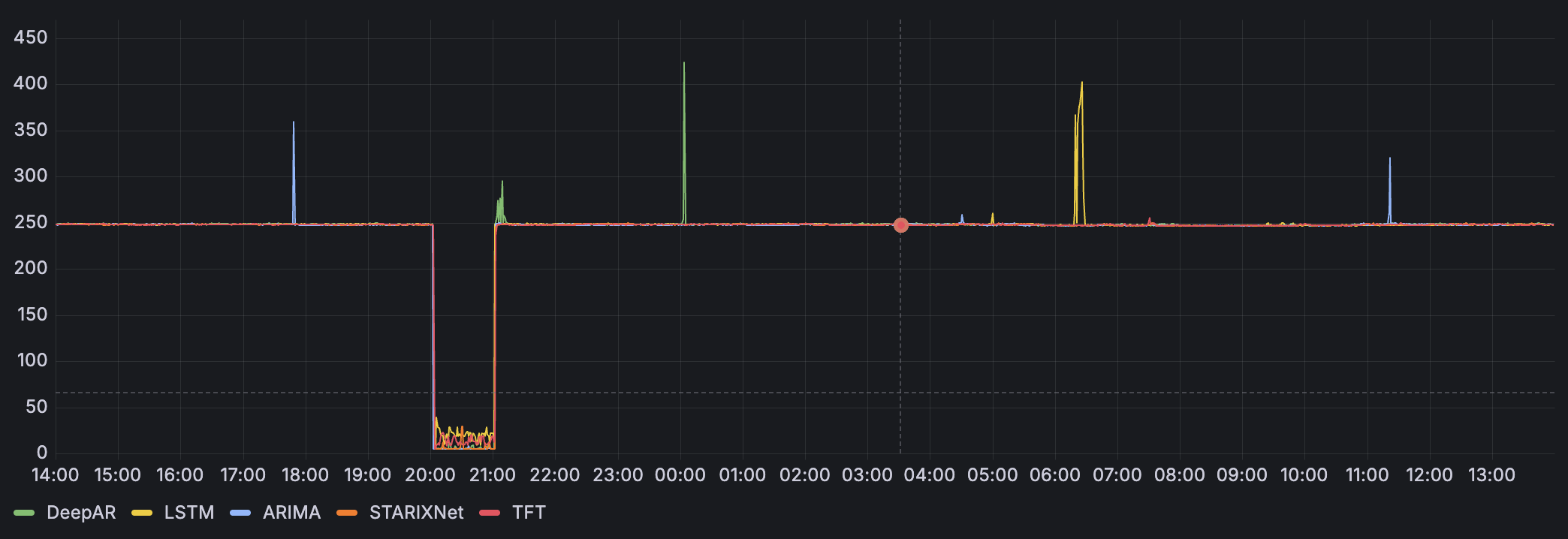}
\caption{Response latency (99th percentile) comparison across microservices during the evaluations. The y-axis shows latency in milliseconds; the x-axis shows the time. Microservices that are scaled by DeepAR, LSTM, and ARIMA models show significant spikes above baseline while STARIXNet and TFT maintain more consistent performance.}
\Description{Response latency (p99) for all microservices during the evaluations.}
\label{fig:latency_p99}
\end{figure}

\begin{figure}[H]
\centering
\includegraphics[width=\linewidth]{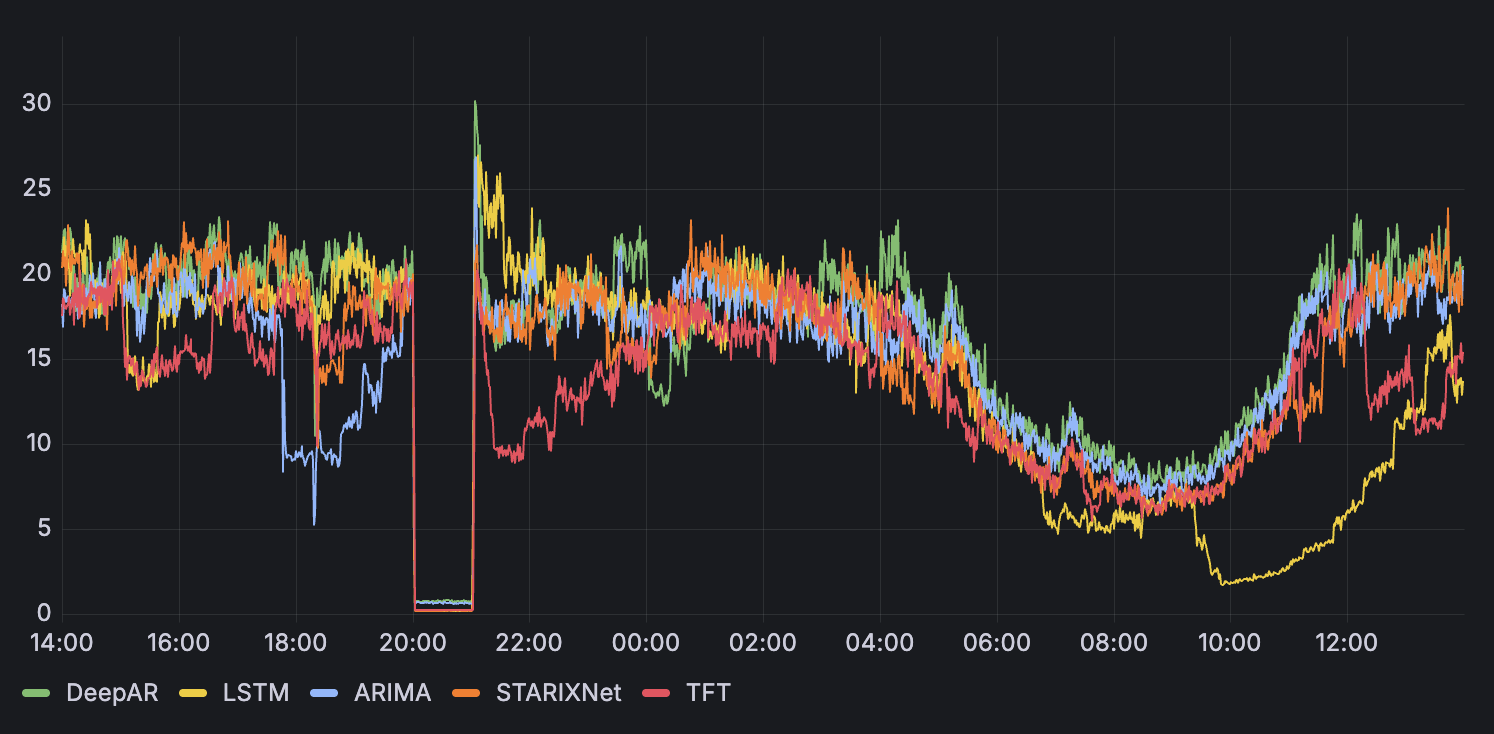}
\caption{CPU usage (50th percentile) comparison under the load during experimentation and evaluations. Each micro service shows distinct resource consumption patterns under identical traffic conditions, because of different scale level. The y-axis shows CPU usage percentage (1-100); the x-axis shows the time. Different microservices shows different CPU usage patterns, with spikes and drops corresponding to load changes and scaling.}
\Description{CPU usage (q50) for all microservices the evaluations.}
\label{fig:cpu_utilization_q50}
\end{figure}

\subsection{Onboarded Client Microservice Figures}\label{ssec:ocmf}

\begin{figure}[H]
  \centering
  \includegraphics[width=\linewidth]{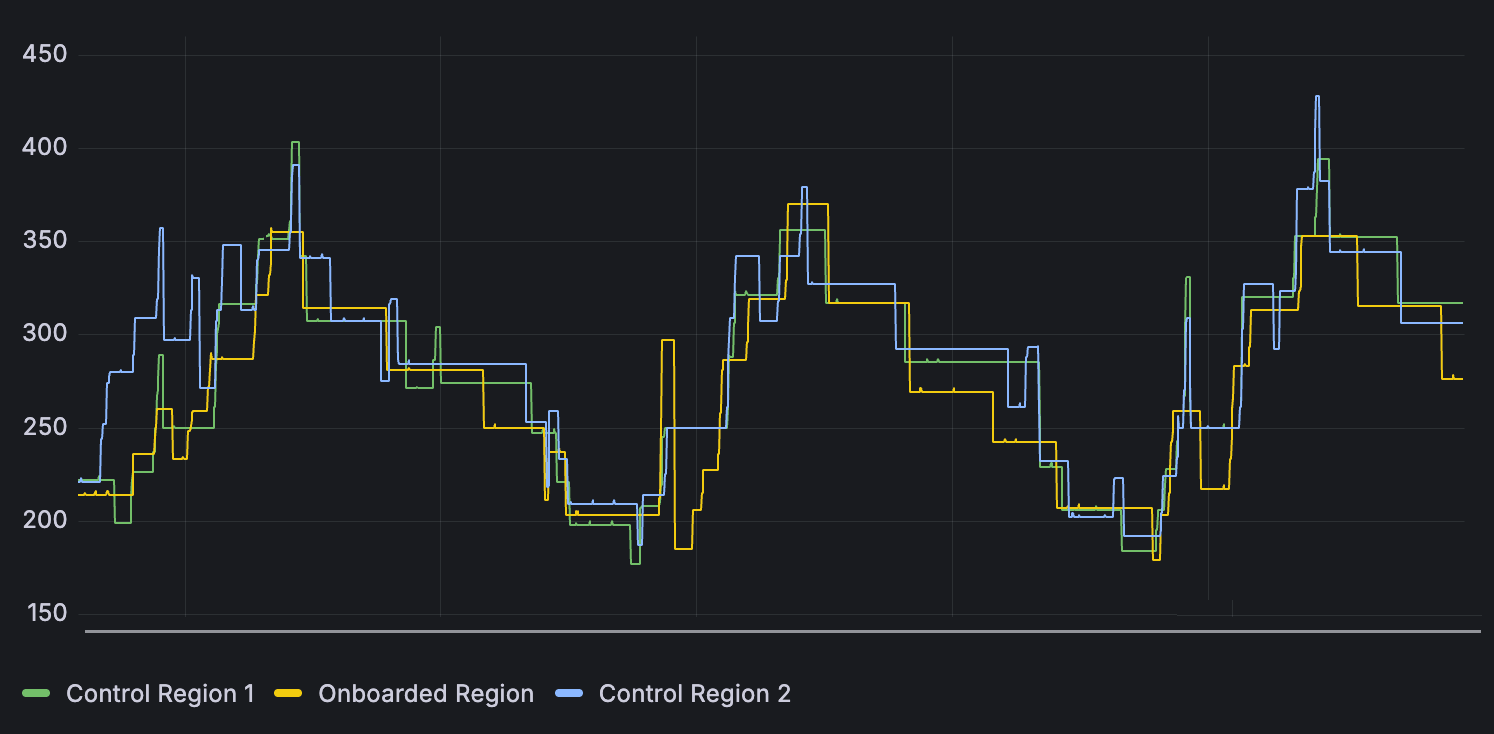}
  \caption{Multi-region comparison post-onboarding a client's microservice on a scale of hundreds of pods. The 2 control regions implement an alternative autoscaling strategy. Plot shows clear distinction in handling volatility, peaks, and predictive scaling in the onboarded region, versus reactive scaling in control regions.}
  \Description{Quasi-A/B testing across regions.}
  \label{fig:post1}
\end{figure}

\begin{figure}[H]
  \centering
  \includegraphics[width=\linewidth]{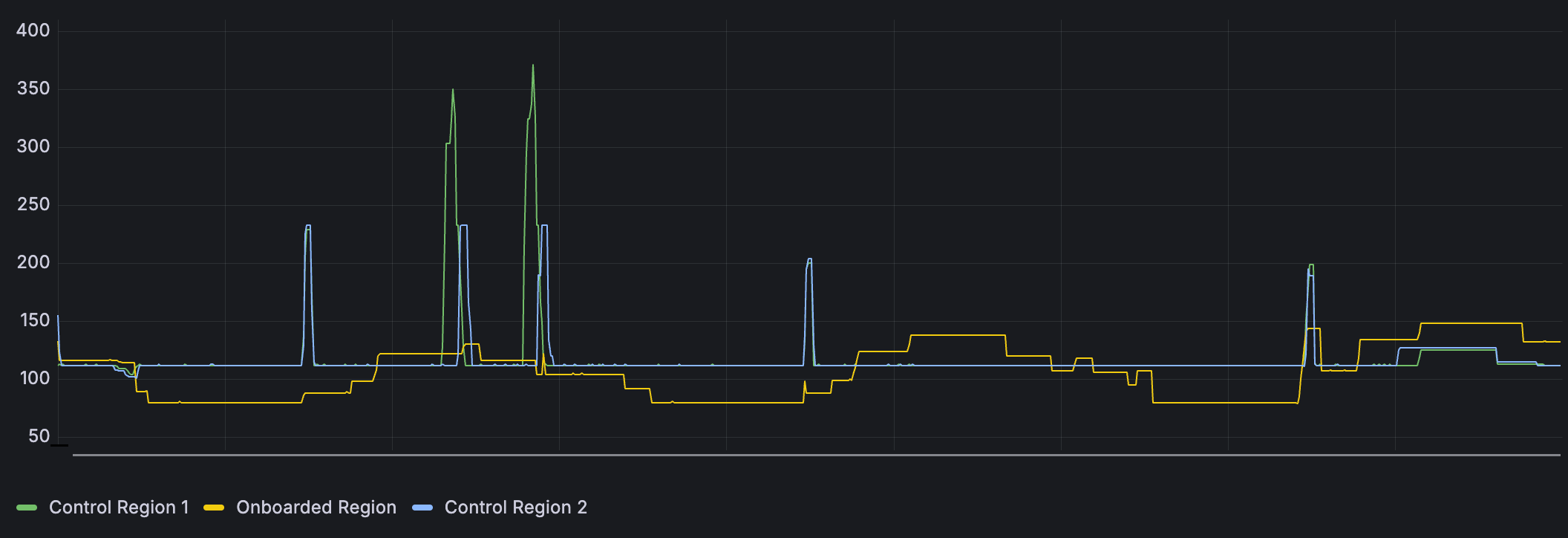}
  \caption{Multi-region comparison post-onboarding of a client’s microservice. The two control regions implement CPU-based scaling with high thresholds for safety, resulting in near-constant capacity regardless of workload spikes or dips. In contrast, the onboarded region adjusts capacity dynamically in response to predicted pod count.}
  \Description{Quasi-A/B testing across regions.}
  \label{fig:post_onboarding_2}
\end{figure}

\begin{figure}[H]
  \centering
  \includegraphics[width=\linewidth]{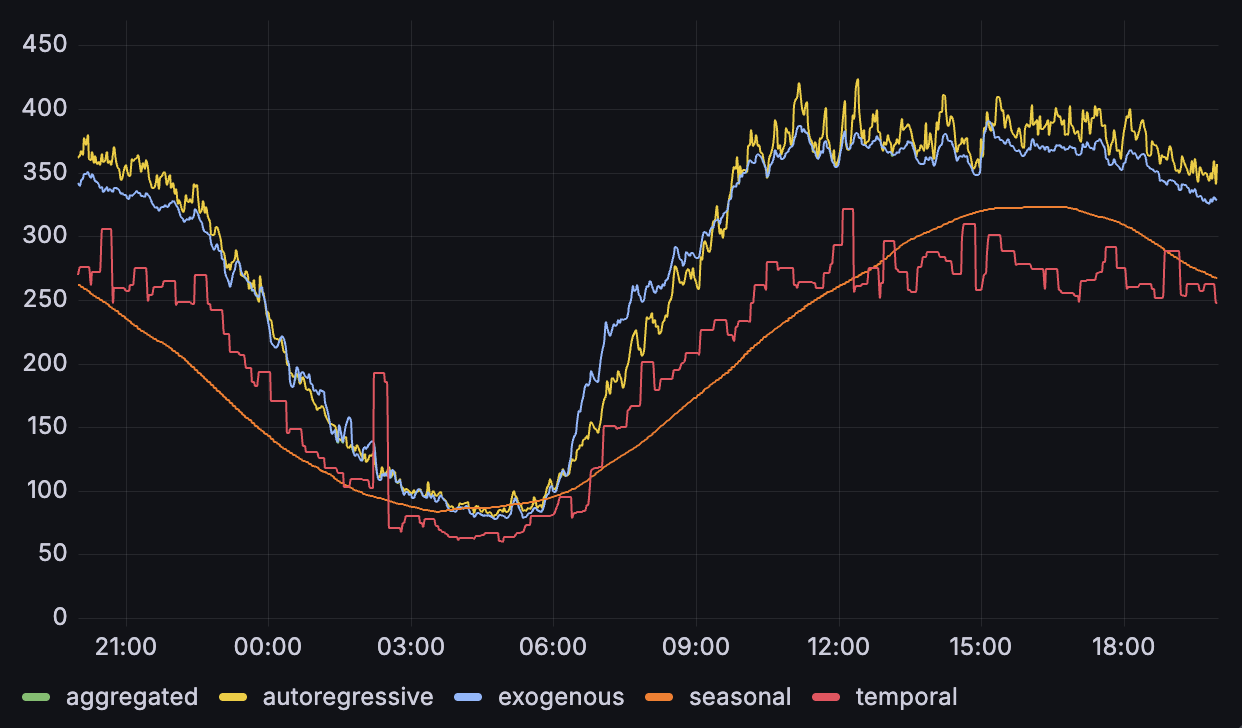}
  \caption{An onboarded microservice's forecasts for the CPU demand (i.e. CPU seconds) from STARIXNet’s decoder modules: autoregressive (yellow), exogenous (blue), seasonal (orange), temporal (red), and aggregated (green).}
  \Description{CPU demand over time time for each decoder output and the final aggregated prediction.}
  \label{fig:decoder_cpu_predictions}
\end{figure}

\begin{figure}[H]
  \centering
  \includegraphics[width=\linewidth]{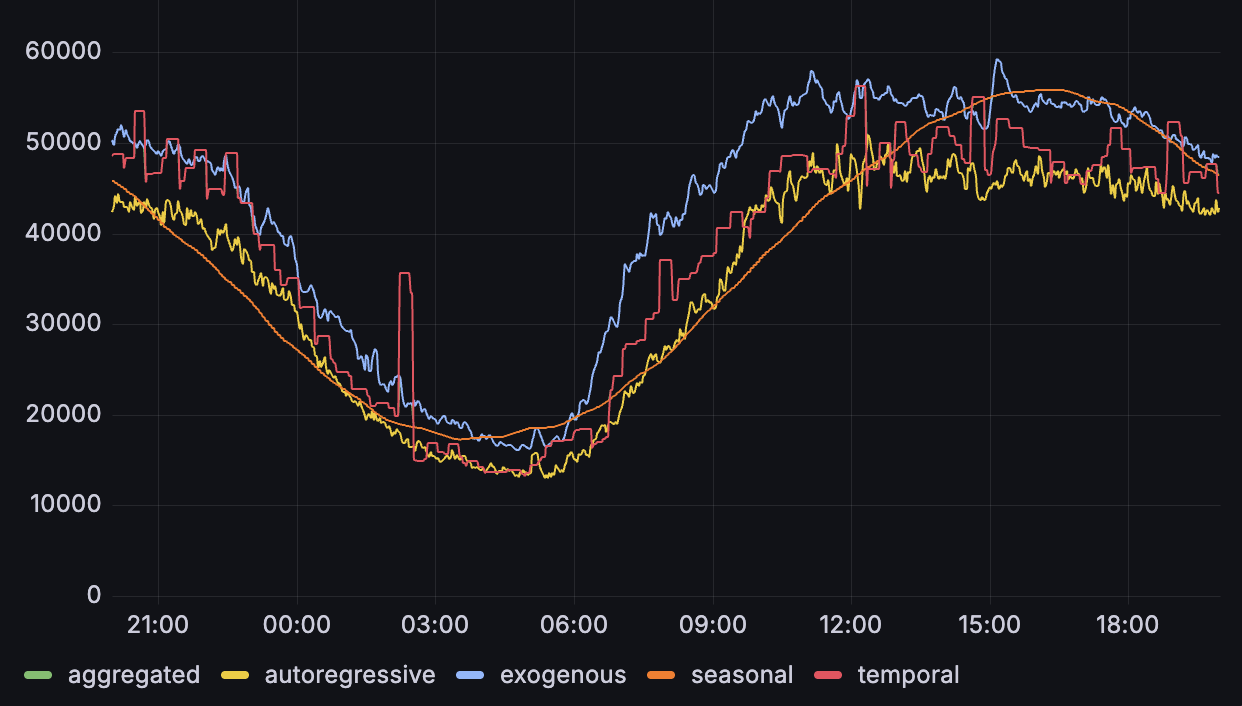}
  \caption{An onboarded microservice's forecasts for the Requests Per Seconds (RPS) from STARIXNet’s decoder modules: autoregressive, exogenous, seasonal, and temporal components.}
  \Description{RPS over time for each decoder output.}
  \label{fig:decoder_tps_predictions}
\end{figure}

\begin{figure}[H]
  \centering
  \includegraphics[width=\linewidth]{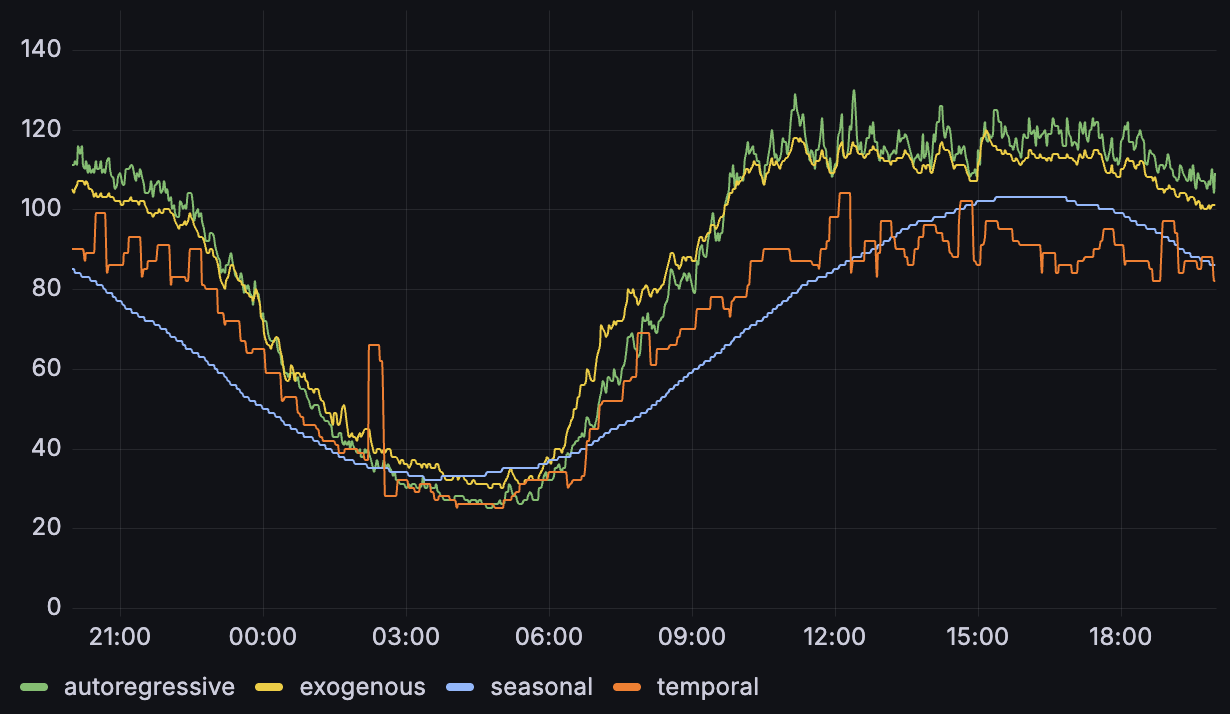}
  \caption{The calculated Pod-count options after aggregating RPS and CPU forecasts from STARIXNet’s decoder modules: autoregressive (green), exogenous (yellow), seasonal (blue), and temporal (orange).}
  \Description{Pod count over time for each decoder output.}
  \label{fig:decoder_pod_predictions}
\end{figure}








\end{document}